\documentclass[11pt]{article}
\usepackage{authblk}
\usepackage{acl2020}
\usepackage{times}
\usepackage{url}
\usepackage{tipa}
\usepackage{graphicx}
\usepackage[T1]{fontenc}
\usepackage{xcolor}
\usepackage{latexsym}
\usepackage{tikz}
\usepackage{adjustbox}

\usepackage[utf8]{inputenc}
\usepackage{pgfplots}
\pgfplotsset{compat=newest}
\usepgfplotslibrary{groupplots}
\usepgfplotslibrary{dateplot}

\aclfinalcopy 

\title{In search of isoglosses: continuous and discrete language embeddings in Slavic historical phonology}

\author[1,2]{\bf Chundra A.\ Cathcart}
\author[3]{\bf Florian Wandl}

\affil[1]{Department of Comparative Language Science, University of Zurich}
\affil[2]{Center for the Interdisciplinary Study of Language Evolution, University of Zurich}
\affil[3]{Slavisches Seminar, University of Zurich}

\affil[ ]{\texttt{\{chundra.cathcart,florian.wandl\}@uzh.ch}}

\date{}

\begin{document}
\maketitle
\begin{abstract}
This paper investigates the ability of neural network architectures to effectively learn diachronic phonological generalizations in a multilingual setting. We employ models using three different types of language embedding (dense, sigmoid, and straight-through). We find that the Straight-Through model outperforms the other two in terms of accuracy, but the Sigmoid model's language embeddings 
show the strongest agreement with the traditional subgrouping of the Slavic languages. 
We find that the Straight-Through model has learned coherent,  semi-interpretable information about sound change, and outline directions for future research. 
\end{abstract}

\section{Introduction}

Historical phonology is an important area of diachronic linguistics, 
allowing scholars to explore the space of possible sound change trajectories and resulting synchronic patterns, as well as posit degrees of relatedness between languages on the basis of sound changes shared across them. 
The latter practice traditionally involves the identification of innovations that are probative with respect to historical subgrouping. The internal genetic structure of many linguistic groups is uncontroversial. 
For others, scholars disagree in terms of which isoglosses are relevant to subgrouping, and whether the relevant features are indeed shared across groups of languages. 
The use of computational methods has aided in resolving a number of outstanding questions in diachronic linguistics, though little work has been done assessing the ability of computational models to learn meaningful patterns of sound change as well as capture language-level information that may bear on degrees of genetic relatedness. 

This paper employs a neural encoder-decoder architecture to analyze patterns of sound change among Slavic languages, training a series of models on data from an etymological dictionary. 
Following the standard practice in multilingual NLP tasks, we make use of language embeddings concatenated to the model input. 
We make use of three different types of language embedding, comprising continuous real-valued {\sc dense}, {\sc sigmoid} (defined on the $[0,1]$ interval), and binary {\sc straight-through} embeddings. 
We assess the accuracy with which these encoder-decoder models predict held-out forms in contemporary Slavic languages from their corresponding Proto-Slavic input. 
We provide a detailed error analysis, observing differences across models in terms of the types of error introduced. 
We measure the extent to which the language embeddings learned by each model recapitulate the 
the most commonly accepted  
subgrouping of the Slavic languages. 
Finally, we assess the interpretability of the straight-through embedding, investigating the degree to which embeddings in binary latent space represent meaningful information regarding sound change. 

We find that the model with straight-through language embeddings 
outperforms the Dense and Sigmoid models in terms of accuracy. 
At the same time, the language embeddings learned by the Sigmoid model display a signal that shows the highest agreement out of the three models with received wisdom regarding the dialect grouping of Slavic languages. 
We find that the latent binary representations learned capture meaningful and coherent information regarding sound patterns. 
We outline future directions for research using latent binary embeddings in neural historical phonology. 


\section{Background}

The Slavic branch of Indo-European is traditionally divided into East, West, and South Slavic groups. 
Many of the oldest and most decisive isoglosses differentiating the Slavic languages are phonological in nature (cf. \citealt{shevelov1964}, \citealt{carlton1991}). 
For instance, tautosyllabic Proto-Slavic vowel$+$liquid sequences were subject to {\sc metathesis} or re-ordering in West and South Slavic languages, whereas East Slavic languages underwent {\sc pleophony}, inserting a vowel between the liquid and the following consonant. 
Variation between liquid metathesis and pleophony, accompanied by language-specific vowel changes, can be seen in the cognates Russian {\it g\'orod}, Ukrainian {\it h\'orod}, Croatian {\it gr\^{a}d}, Czech {\it hrad} ($<$ {*g\^ord\u{u}} `city'); the {\it h-} found in Ukrainian (East Slavic) and Czech (West Slavic) show also that certain shared features do not cleanly follow the taxonomy defined above. 

It is traditionally assumed that the tripartite classification of Slavic either reflects the dialectal diversity of the so-called Slavic homeland, most probably situated on the outskirts of the Carpathian Mountains, or emerged as a result of the great Slavic expansion in the 6th century AD \citep{brauer1961,hock1998}. The extensive study of loanwords, however, suggests that post-expansional Slavic was, despite the vast territory it occupied, still uniform. There seem to have been no significant differences between Slavic spoken in areas located as far away from each other as the Baltic sea and the Peloponnese, at least with regard to phonology. It has therefore been argued that it is this post-expansional Slavic that constitutes the ancestor of all Slavic languages and not the Slavic language spoken in the homeland \citep{holzer1995}. One of the arguments put forward in support of this claim is the still largely reconstructible post-Proto-Slavic dialect continuum \citep{holzer1997}. 
One objective of this paper to assess the degree to which neural models recapitulate the uncontroversial subgrouping of Slavic as an indicator of whether they are capable of resolving outstanding issues in the field.


\section{Related Work}
A growing body of research assesses the information captured by language embeddings trained on large data sets using neural models. 
There is some debate as to whether embeddings learned in these tasks can pick up on genetic signal \citep{OstlingTiedemann2017,Tiedemann2018}, or whether the information learned represents structural similarity \citep{Bjervaetal2019}. 
The majority of work of language embeddings involves models trained on large parallel corpora. 
\citet{Meloni2019} approach the issue of sound change using a GRU-based neural machine translation model with soft attention to reconstruct Latin forms from contemporary Romance reflexes; the authors employ language embeddings, but do not provide an analysis of the information captured by these embeddings. 
Phylogenetic approaches to sound change and the reconstruction of word forms incorporate a highly articulated genetic representation of language relatedness \citep{Hruschkaetal2013,BouchardCote2013}, but employ simplified representations of sound change in comparison to what can be captured by recurrent neural networks; at the same time, phylogenetic work explicitly models intermediate stages of change, a potential challenge for RNNs, which are better suited to learning patterns resulting from the telescoping of multiple changes. Related work seeks to disentangle genetic and areal pressures in shaping cross-linguistic patterns  \citep{Daume2009,murawaki2018statistical,cathcart2019toward,cathcart2019probabilistic,cathcart2020dialectal}. 

In general, while the signal learned by embeddings can be analyzed via visualization techniques \citep{maaten2008visualizing}, it is a challenge to link the behavior of embeddings to individual features in the data analyzed. 
This difficulty undoubtedly stems in part from the fact that embeddings are generally continuous, lacking the sparsity or discreteness needed to identify the behavior of the neural model when features are active or inactive. 
This issue has been addressed by the development of de-noising approaches designed to induce sparsity \citep{subramanian2018spine}. 

Binary latent variables are of key interest to linguistic questions, but pose many challenges for inference. 
Binary latent variable models such as the Indian Buffet Process \citep[IBP, ][]{Ghahramani2006} have been used in some applications in computational phonology and typology \cite{Doyleetal2014,Murawaki2017} using a combination of Gibbs Sampling and updates from the Metropolis-Hastings algorithm or Hamiltonian Monte Carlo, but it is not clear that these inference procedures are scalable to neural models. 
Discrete variables pose problems for differentiability in gradient-based optimization algorithms; marginalizing out all possible combinations of binary variables is generally unfeasible for binary latent variables. 
Variational approaches have attempted to circumvent this issue via the concrete (alternatively, Gumbel-Softmax) distribution \citep{maddison2016concrete,jang2016categorical}, which extends the reparameterization trick to categorical distributions and which produces gradient estimates that have lower variance than standard estimation techniques \cite{williams1992simple} but are still biased; subsequent approaches reduce bias but are less straightforward to implement  \citep{grathwohl2017backpropagation,liu2018rao}. 

While concrete-distributed versions of the IBP have been used in neural models \citep{singh2017structured,kessler2019indian}, this work is limited to variational autoencoders, which use amortized variational inference to learn latent representations from the data via a global inference network; encoder-decoder mechanisms with attention like the one used in this paper cannot exploit this property of the data; training the latent variable with stochastic variational inference, while theoretically possible, is considerably more difficult \citep{kim2018tutorial}. 
As an alternative, we use straight-through (ST) embeddings \citep{bengio2013estimating,courbariaux2016binarized} in a maximum likelihood framework. Straight-through layers are discrete but have underlying continuous weights; model output is predicted on the basis of the discrete representation, while model loss is differentiated with respect to the continuous underlying weights. While this approach has the same problems with biased estimates as the concrete distribution, it is straightforward to implement. We compare the quality of straight-through embeddings to embeddings with no activation and embeddings with sigmoid activation. 

\section{Data}

Our data set consists of Proto-Slavic etyma and corresponding reflexes in medieval and modern Slavic languages taken from a digitized version of a Slavic etymological dictionary (\citealt{derksen2007etymological}; for alternative reconstructions see \citealt{holzer1995,andersen1998}). 
In order to minimize the chance of introducing morphologically non-congruent forms into our data set, we extracted the first form provided for each Slavic language in each entry, since these are most likely to agree morphologically with the Proto-Slavic headword. 

We converted forms in modern Slavic languages to a narrow phonetic representation using IPA transcriptions from Wiktionary (\url{https://www.wiktionary.org}), which were used to train a neural encoder-decoder; these models were used to obtain IPA transcriptions for forms not in Wiktionary, and a portion was checked manually. In several cases we reconciled sources used in the etymological dictionary  \citep[e.g.,][]{pletersnik1894} with contemporary standardized orthographies, and made use of phonetic descriptions for languages where the training data were problematic \citep{schustersewc1968,lencek1982,scatton1984,comriecorbett1993,ternesvladimirova1990,landauetal1995,sustarsicetal1995,dankovicova1997,jassem2003,gussmann2007,stadnik2009,hanulikovahamann2010,mojsijenko2010,yanushevskayabuncic2015,howson2017,howson2018,pompinoetal2017}.
For the medieval languages Old Church Slavic and Church Slavic, orthographic forms were converted to a broad phonemic transcription based on \citet{lunt2001}. Suprasegmental features were marked for all modern languages (pitch accent for Slovene and BCS and primary stress for the remainder; for consistency, we chose to mark primary stress on monosyllables in stress-timed languages). 
We excluded languages with fewer than 100 forms in the etymological dictionary (this resulted in the omission of Macedonian, Polabian and Slovincian). 

We took additional steps to remove morphological mismatches in the data set. 
For Bulgarian verbs, which reflect the Proto-Slavic 1sg present in their citation form, we replaced the Proto-Slavic headword (the infinitive form by default) with a morphologically congruent form, and excluded a small number of forms based on athematic verbs. 
Additionally, Proto-Slavic adjectives are always given in the nominal or short form, although contemporary Slavic languages often reflect the so-called long form, which arose from the addition of an inflected element {*-j\u{\i}} to the ending; 
we converted short Proto-Slavic adjectives to their long form in the appropriate contexts. 
We tried to ensure that Proto-Slavic verbs matched their reflexes according to the presence/absence of reflexive morphology and preverbs. 
Additionally, the original data source contains multiple gender inflections for certain Proto-Slavic etyma (e.g., {*\`abl\u{u}ko} n., {*\`abl\u{u}ka} f., and {*\`abl\u{u}k\u{u}} m.\  for `apple'), which are linked to the same reflexes irrespective of the reflexes' gender; for such forms, we discarded etymon-reflex pairs with mismatched gender. 
Ultimately, this process yielded 11400 forms in 13 languages (see Table \ref{tab:lang_counts}), and allowed us to rid the data set of a large number (albeit not the entirety) of morphological mismatches. 






\begin{table}
\small{
    \centering
    \begin{tabular}{|l|l|c|}
\hline
Language & Glottocode & \# reflexes \\
\hline
\hline
Russian (Rus) & russ1263 & 1572 \\
Slovene (Sln) & slov1268 & 1462 \\
Serbo-Croatian (BCS[M]) & sout1528 & 1434 \\
Czech (Cze) & czec1258 & 1377 \\
Polish (Pol) & poli1260 & 1282 \\
Slovak (Slk) & slov1269 & 1091 \\
Old Church Slavic (OCS) & chur1257 & 1097 \\
Bulgarian (Bul) & bulg1262  & 950 \\
Church Slavic (CS) & chur1257 & 392 \\
Ukrainian (Ukr) & ukra1253  & 301 \\
Upper Sorbian (USo) & uppe1395 & 243 \\
Lower Sorbian (LSo) & lowe1385 & 120 \\
Belarusian (Bel) & bela1254 & 79 \\
\hline
Total &  & 11400\\
\hline
    \end{tabular}
    \caption{Number of forms in each language in data set, along with closest matching glottocodes.}
    \label{tab:lang_counts}
    }
\end{table}

\section{Method}

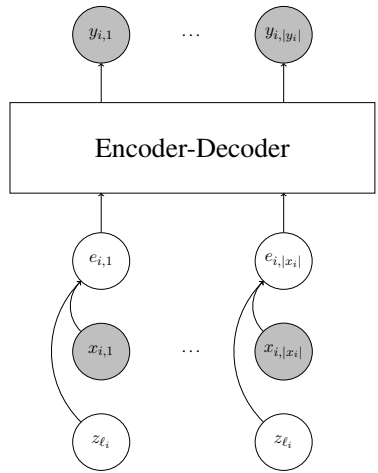
\begin{figure}

\begin{adjustbox}{max totalsize={.8\linewidth}{.8\linewidth},center}
\begin{tikzpicture}
\centering
\draw (0,-2) node[circle,minimum size=1.25cm,draw] (z0) {$z_{\ell_i}$};
\draw (4,-2) node[circle,minimum size=1.25cm,draw] (z1) {$z_{\ell_i}$};
\draw (0,0) node[circle,minimum size=1.25cm,fill=lightgray,draw] (x0) {$x_{i,1}$};
\draw (4,0) node[circle,minimum size=1.25cm,fill=lightgray,draw] (x1) {$x_{i,|x_i|}$};
\draw (0,2) node[circle,minimum size=1.25cm,draw] (e0) {$e_{i,1}$};
\draw (4,2) node[circle,minimum size=1.25cm,draw] (e1) {$e_{i,|x_i|}$};

\draw[draw=black] (-2,3.5) rectangle ++(8,2) node[pos=.5] (box) {\LARGE{Encoder-Decoder}};

\draw (0,7) node[circle,minimum size=1.25cm,fill=lightgray,draw] (y0) {$y_{i,1}$};
\draw (4,7) node[circle,minimum size=1.25cm,fill=lightgray,draw] (y1) {$y_{i,|y_i|}$};

\draw [->] (z0) to [bend left=45] (e0);
\draw [->] (z1) to [bend left=45] (e1);

\draw [->] (x0) to [bend left=45] (e0);
\draw [->] (x1) to [bend left=45] (e1);

\draw [->] (e0) to (0,3.5);
\draw [->] (e1) to (4,3.5);

\draw [->] (0,5.5) to (y0);
\draw [->] (4,5.5) to (y1);

\draw (2,0) node {\ldots};
\draw (2,7) node {\ldots};
\end{tikzpicture}
\end{adjustbox}

\caption{Basic schema of architecture used in this paper; for each input-output pair $\mathbf x_i,\mathbf y_i$, an embedding associated with the language ID for index $i$ is concatenated to a one-hot representation of the input.}
\label{architecture}
\end{figure}

To learn mappings between Proto-Slavic etyma and the Slavic reflexes that descend from them, we use an LSTM Encoder-Decoder with 0th-order hard monotonic attention \citep{wu-cotterell-2019-exact}, trained on all languages in our data set. 
The basic model architecture used for the experiments in this study has the following structure (schematized in Figure \ref{architecture}): a trainable language-level embedding is concatenated to a one-hot representation of each input segment at each input time step; each concatenation is fed to a Dense layer (with no activation) to generate a embedding for each time step that encodes information about the input phoneme and language ID of the reflex; these embeddings subsequently are fed to the encoder-decoder in order to generate the output. 
The parameters of the encoder-decoder architecture are shared across languages in the data set; the sole language-specific variable employed is the language-level embedding fed to the model.

In all experiments, we set the dimension of the language-level embedding and the language/character embedding to 128, and the hidden layer dimension to 256. 
In our experiments, we employ different representations of the language-level embedding, including a dense layer with no activation ({\sc dense} model), a dense layer with sigmoid activation, ({\sc sigmoid} model) and a dense layer with straight-through activation ({\sc ST} model), which uses the Heaviside step function (negative values map to 0, non-negative values to 1). 
We train our model for 200 epochs with a batch size of 256 using the Adam optimizer with a learning rate of $.001$ with the objective of minimizing the mean categorical cross-entropy between the predicted and observed distributions of the output. To evaluate model performance, we carry out $K$-fold cross-validation ($K=10$), randomly holding out 10\% of the forms in each language, and greedily decoding the held-out forms using the trained model. For additional analyses regarding the interpretability of the embeddings learned, we train the model on all forms in the data set. 
Models are implemented in Keras \citep{chollet2015} and Larq \citep{larq}.\footnote{Code accompanying this paper is available at \url{https://github.com/chundrac/slav-dial/tree/master/SIGMORPHON_2020}}

\section{Results}
\subsection{Accuracy}

\begin{figure*}
    \centering
    \begin{minipage}[t]{.3\linewidth}
    \adjustbox{width=\linewidth}{
\begin{tikzpicture}

\definecolor{color0}{rgb}{0.12156862745098,0.466666666666667,0.705882352941177}

\begin{axis}[
tick align=outside,
tick pos=left,
x grid style={white!69.0196078431373!black},
xmin=4.21991504644092, xmax=7.50963677901526,
xtick style={color=black},
y grid style={white!69.0196078431373!black},
ymin=0.0617795091487563, ymax=0.294028607122557,
ytick style={color=black}
]
\addplot [semithick, color0, opacity=0, mark=*, mark size=3, mark options={solid}, only marks]
table {%
7.26822302115957 0.151291519963068
};
\addplot [semithick, color0, opacity=0, mark=*, mark size=3, mark options={solid}, only marks]
table {%
4.36944785246702 0.233322654841642
};
\addplot [semithick, color0, opacity=0, mark=*, mark size=3, mark options={solid}, only marks]
table {%
6.85646198459459 0.186557909926331
};
\addplot [semithick, color0, opacity=0, mark=*, mark size=3, mark options={solid}, only marks]
table {%
5.97126183979046 0.0723362863293836
};
\addplot [semithick, color0, opacity=0, mark=*, mark size=3, mark options={solid}, only marks]
table {%
7.22766249872865 0.135496335093341
};
\addplot [semithick, color0, opacity=0, mark=*, mark size=3, mark options={solid}, only marks]
table {%
4.78749174278205 0.214311683686684
};
\addplot [semithick, color0, opacity=0, mark=*, mark size=3, mark options={solid}, only marks]
table {%
7.00033446027523 0.0785844308633734
};
\addplot [semithick, color0, opacity=0, mark=*, mark size=3, mark options={solid}, only marks]
table {%
7.15617663748062 0.111032899661047
};
\addplot [semithick, color0, opacity=0, mark=*, mark size=3, mark options={solid}, only marks]
table {%
7.36010397298915 0.158515321190725
};
\addplot [semithick, color0, opacity=0, mark=*, mark size=3, mark options={solid}, only marks]
table {%
6.99484998583307 0.12867719153237
};
\addplot [semithick, color0, opacity=0, mark=*, mark size=3, mark options={solid}, only marks]
table {%
7.28756064030972 0.161902195691525
};
\addplot [semithick, color0, opacity=0, mark=*, mark size=3, mark options={solid}, only marks]
table {%
5.70711026474888 0.28347182994193
};
\addplot [semithick, color0, opacity=0, mark=*, mark size=3, mark options={solid}, only marks]
table {%
5.49306144334055 0.190895312963214
};
\draw (axis cs:7.26822302115957,0.151291519963068) node[
  scale=1.2,
  anchor=base west,
  text=black,
  rotate=0.0
]{BCS};
\draw (axis cs:4.36944785246702,0.233322654841642) node[
  scale=1.2,
  anchor=base west,
  text=black,
  rotate=0.0
]{Bel};
\draw (axis cs:6.85646198459459,0.186557909926331) node[
  scale=1.2,
  anchor=base west,
  text=black,
  rotate=0.0
]{Bul};
\draw (axis cs:5.97126183979046,0.0723362863293836) node[
  scale=1.2,
  anchor=base west,
  text=black,
  rotate=0.0
]{CS};
\draw (axis cs:7.22766249872865,0.135496335093341) node[
  scale=1.2,
  anchor=base west,
  text=black,
  rotate=0.0
]{Cze};
\draw (axis cs:4.78749174278205,0.214311683686684) node[
  scale=1.2,
  anchor=base west,
  text=black,
  rotate=0.0
]{LSo};
\draw (axis cs:7.00033446027523,0.0785844308633734) node[
  scale=1.2,
  anchor=base west,
  text=black,
  rotate=0.0
]{OCS};
\draw (axis cs:7.15617663748062,0.111032899661047) node[
  scale=1.2,
  anchor=base west,
  text=black,
  rotate=0.0
]{Pol};
\draw (axis cs:7.36010397298915,0.158515321190725) node[
  scale=1.2,
  anchor=base west,
  text=black,
  rotate=0.0
]{Rus};
\draw (axis cs:6.99484998583307,0.12867719153237) node[
  scale=1.2,
  anchor=base west,
  text=black,
  rotate=0.0
]{Slk};
\draw (axis cs:7.28756064030972,0.161902195691525) node[
  scale=1.2,
  anchor=base west,
  text=black,
  rotate=0.0
]{Sln};
\draw (axis cs:5.70711026474888,0.28347182994193) node[
  scale=1.2,
  anchor=base west,
  text=black,
  rotate=0.0
]{Ukr};
\draw (axis cs:5.49306144334055,0.190895312963214) node[
  scale=1.2,
  anchor=base west,
  text=black,
  rotate=0.0
]{USo};
\end{axis}

\end{tikzpicture}
    }
    \end{minipage}
    \hspace{.03\linewidth}
    \begin{minipage}[t]{.3\linewidth}
    \adjustbox{width=\linewidth}{
\begin{tikzpicture}

\definecolor{color0}{rgb}{0.12156862745098,0.466666666666667,0.705882352941177}

\begin{axis}[
tick align=outside,
tick pos=left,
x grid style={white!69.0196078431373!black},
xmin=4.21991504644092, xmax=7.50963677901526,
xtick style={color=black},
y grid style={white!69.0196078431373!black},
ymin=0.063657736272397, ymax=0.320328242546719,
ytick style={color=black}
]
\addplot [semithick, color0, opacity=0, mark=*, mark size=3, mark options={solid}, only marks]
table {%
7.26822302115957 0.159283880152081
};
\addplot [semithick, color0, opacity=0, mark=*, mark size=3, mark options={solid}, only marks]
table {%
4.36944785246702 0.226621058266628
};
\addplot [semithick, color0, opacity=0, mark=*, mark size=3, mark options={solid}, only marks]
table {%
6.85646198459459 0.18727665317139
};
\addplot [semithick, color0, opacity=0, mark=*, mark size=3, mark options={solid}, only marks]
table {%
5.97126183979046 0.0753245774666843
};
\addplot [semithick, color0, opacity=0, mark=*, mark size=3, mark options={solid}, only marks]
table {%
7.22766249872865 0.141412859314071
};
\addplot [semithick, color0, opacity=0, mark=*, mark size=3, mark options={solid}, only marks]
table {%
4.78749174278205 0.215433224183224
};
\addplot [semithick, color0, opacity=0, mark=*, mark size=3, mark options={solid}, only marks]
table {%
7.00033446027523 0.0863036321585212
};
\addplot [semithick, color0, opacity=0, mark=*, mark size=3, mark options={solid}, only marks]
table {%
7.15617663748062 0.115706063843588
};
\addplot [semithick, color0, opacity=0, mark=*, mark size=3, mark options={solid}, only marks]
table {%
7.36010397298915 0.166628452072716
};
\addplot [semithick, color0, opacity=0, mark=*, mark size=3, mark options={solid}, only marks]
table {%
6.99484998583307 0.137237940285603
};
\addplot [semithick, color0, opacity=0, mark=*, mark size=3, mark options={solid}, only marks]
table {%
7.28756064030972 0.171019980517949
};
\addplot [semithick, color0, opacity=0, mark=*, mark size=3, mark options={solid}, only marks]
table {%
5.70711026474888 0.308661401352431
};
\addplot [semithick, color0, opacity=0, mark=*, mark size=3, mark options={solid}, only marks]
table {%
5.49306144334055 0.207756812232121
};
\draw (axis cs:7.26822302115957,0.159283880152081) node[
  scale=1.2,
  anchor=base west,
  text=black,
  rotate=0.0
]{BCS};
\draw (axis cs:4.36944785246702,0.226621058266628) node[
  scale=1.2,
  anchor=base west,
  text=black,
  rotate=0.0
]{Bel};
\draw (axis cs:6.85646198459459,0.18727665317139) node[
  scale=1.2,
  anchor=base west,
  text=black,
  rotate=0.0
]{Bul};
\draw (axis cs:5.97126183979046,0.0753245774666843) node[
  scale=1.2,
  anchor=base west,
  text=black,
  rotate=0.0
]{CS};
\draw (axis cs:7.22766249872865,0.141412859314071) node[
  scale=1.2,
  anchor=base west,
  text=black,
  rotate=0.0
]{Cze};
\draw (axis cs:4.78749174278205,0.215433224183224) node[
  scale=1.2,
  anchor=base west,
  text=black,
  rotate=0.0
]{LSo};
\draw (axis cs:7.00033446027523,0.0863036321585212) node[
  scale=1.2,
  anchor=base west,
  text=black,
  rotate=0.0
]{OCS};
\draw (axis cs:7.15617663748062,0.115706063843588) node[
  scale=1.2,
  anchor=base west,
  text=black,
  rotate=0.0
]{Pol};
\draw (axis cs:7.36010397298915,0.166628452072716) node[
  scale=1.2,
  anchor=base west,
  text=black,
  rotate=0.0
]{Rus};
\draw (axis cs:6.99484998583307,0.137237940285603) node[
  scale=1.2,
  anchor=base west,
  text=black,
  rotate=0.0
]{Slk};
\draw (axis cs:7.28756064030972,0.171019980517949) node[
  scale=1.2,
  anchor=base west,
  text=black,
  rotate=0.0
]{Sln};
\draw (axis cs:5.70711026474888,0.308661401352431) node[
  scale=1.2,
  anchor=base west,
  text=black,
  rotate=0.0
]{Ukr};
\draw (axis cs:5.49306144334055,0.207756812232121) node[
  scale=1.2,
  anchor=base west,
  text=black,
  rotate=0.0
]{USo};
\end{axis}

\end{tikzpicture}
    }
    \end{minipage}
    \hspace{.03\linewidth}
    \begin{minipage}[t]{.3\linewidth}
    \adjustbox{width=\linewidth}{
\begin{tikzpicture}

\definecolor{color0}{rgb}{0.12156862745098,0.466666666666667,0.705882352941177}

\begin{axis}[
tick align=outside,
tick pos=left,
x grid style={white!69.0196078431373!black},
xmin=4.21991504644092, xmax=7.50963677901526,
xtick style={color=black},
y grid style={white!69.0196078431373!black},
ymin=0.054918593877487, ymax=0.285105170712412,
ytick style={color=black}
]
\addplot [semithick, color0, opacity=0, mark=*, mark size=3, mark options={solid}, only marks]
table {%
7.26822302115957 0.146460089190215
};
\addplot [semithick, color0, opacity=0, mark=*, mark size=3, mark options={solid}, only marks]
table {%
4.36944785246702 0.206023968998653
};
\addplot [semithick, color0, opacity=0, mark=*, mark size=3, mark options={solid}, only marks]
table {%
6.85646198459459 0.180204546915073
};
\addplot [semithick, color0, opacity=0, mark=*, mark size=3, mark options={solid}, only marks]
table {%
5.97126183979046 0.0653816200972563
};
\addplot [semithick, color0, opacity=0, mark=*, mark size=3, mark options={solid}, only marks]
table {%
7.22766249872865 0.136055337282778
};
\addplot [semithick, color0, opacity=0, mark=*, mark size=3, mark options={solid}, only marks]
table {%
4.78749174278205 0.18610421985422
};
\addplot [semithick, color0, opacity=0, mark=*, mark size=3, mark options={solid}, only marks]
table {%
7.00033446027523 0.0759042638714471
};
\addplot [semithick, color0, opacity=0, mark=*, mark size=3, mark options={solid}, only marks]
table {%
7.15617663748062 0.107877233602204
};
\addplot [semithick, color0, opacity=0, mark=*, mark size=3, mark options={solid}, only marks]
table {%
7.36010397298915 0.152664514591996
};
\addplot [semithick, color0, opacity=0, mark=*, mark size=3, mark options={solid}, only marks]
table {%
6.99484998583307 0.129830768579623
};
\addplot [semithick, color0, opacity=0, mark=*, mark size=3, mark options={solid}, only marks]
table {%
7.28756064030972 0.163633689232184
};
\addplot [semithick, color0, opacity=0, mark=*, mark size=3, mark options={solid}, only marks]
table {%
5.70711026474888 0.274642144492643
};
\addplot [semithick, color0, opacity=0, mark=*, mark size=3, mark options={solid}, only marks]
table {%
5.49306144334055 0.188809230321576
};
\draw (axis cs:7.26822302115957,0.146460089190215) node[
  scale=1.2,
  anchor=base west,
  text=black,
  rotate=0.0
]{BCS};
\draw (axis cs:4.36944785246702,0.206023968998653) node[
  scale=1.2,
  anchor=base west,
  text=black,
  rotate=0.0
]{Bel};
\draw (axis cs:6.85646198459459,0.180204546915073) node[
  scale=1.2,
  anchor=base west,
  text=black,
  rotate=0.0
]{Bul};
\draw (axis cs:5.97126183979046,0.0653816200972563) node[
  scale=1.2,
  anchor=base west,
  text=black,
  rotate=0.0
]{CS};
\draw (axis cs:7.22766249872865,0.136055337282778) node[
  scale=1.2,
  anchor=base west,
  text=black,
  rotate=0.0
]{Cze};
\draw (axis cs:4.78749174278205,0.18610421985422) node[
  scale=1.2,
  anchor=base west,
  text=black,
  rotate=0.0
]{LSo};
\draw (axis cs:7.00033446027523,0.0759042638714471) node[
  scale=1.2,
  anchor=base west,
  text=black,
  rotate=0.0
]{OCS};
\draw (axis cs:7.15617663748062,0.107877233602204) node[
  scale=1.2,
  anchor=base west,
  text=black,
  rotate=0.0
]{Pol};
\draw (axis cs:7.36010397298915,0.152664514591996) node[
  scale=1.2,
  anchor=base west,
  text=black,
  rotate=0.0
]{Rus};
\draw (axis cs:6.99484998583307,0.129830768579623) node[
  scale=1.2,
  anchor=base west,
  text=black,
  rotate=0.0
]{Slk};
\draw (axis cs:7.28756064030972,0.163633689232184) node[
  scale=1.2,
  anchor=base west,
  text=black,
  rotate=0.0
]{Sln};
\draw (axis cs:5.70711026474888,0.274642144492643) node[
  scale=1.2,
  anchor=base west,
  text=black,
  rotate=0.0
]{Ukr};
\draw (axis cs:5.49306144334055,0.188809230321576) node[
  scale=1.2,
  anchor=base west,
  text=black,
  rotate=0.0
]{USo};
\end{axis}

\end{tikzpicture}
    }
    \end{minipage}
    \vspace{.03\linewidth}
    \begin{minipage}[t]{.3\linewidth}
    \adjustbox{width=\linewidth}{
\begin{tikzpicture}

\definecolor{color0}{rgb}{0.12156862745098,0.466666666666667,0.705882352941177}

\begin{axis}[
tick align=outside,
tick pos=left,
x grid style={white!69.0196078431373!black},
xmin=4.21991504644092, xmax=7.50963677901526,
xtick style={color=black},
y grid style={white!69.0196078431373!black},
ymin=0.299314784053156, ymax=0.841940555291884,
ytick style={color=black}
]
\addplot [semithick, color0, opacity=0, mark=*, mark size=3, mark options={solid}, only marks]
table {%
7.26822302115957 0.583682008368201
};
\addplot [semithick, color0, opacity=0, mark=*, mark size=3, mark options={solid}, only marks]
table {%
4.36944785246702 0.759493670886076
};
\addplot [semithick, color0, opacity=0, mark=*, mark size=3, mark options={solid}, only marks]
table {%
6.85646198459459 0.594736842105263
};
\addplot [semithick, color0, opacity=0, mark=*, mark size=3, mark options={solid}, only marks]
table {%
5.97126183979046 0.323979591836735
};
\addplot [semithick, color0, opacity=0, mark=*, mark size=3, mark options={solid}, only marks]
table {%
7.22766249872865 0.565722585330429
};
\addplot [semithick, color0, opacity=0, mark=*, mark size=3, mark options={solid}, only marks]
table {%
4.78749174278205 0.75
};
\addplot [semithick, color0, opacity=0, mark=*, mark size=3, mark options={solid}, only marks]
table {%
7.00033446027523 0.333637192342753
};
\addplot [semithick, color0, opacity=0, mark=*, mark size=3, mark options={solid}, only marks]
table {%
7.15617663748062 0.497659906396256
};
\addplot [semithick, color0, opacity=0, mark=*, mark size=3, mark options={solid}, only marks]
table {%
7.36010397298915 0.513358778625954
};
\addplot [semithick, color0, opacity=0, mark=*, mark size=3, mark options={solid}, only marks]
table {%
6.99484998583307 0.575618698441796
};
\addplot [semithick, color0, opacity=0, mark=*, mark size=3, mark options={solid}, only marks]
table {%
7.28756064030972 0.536935704514364
};
\addplot [semithick, color0, opacity=0, mark=*, mark size=3, mark options={solid}, only marks]
table {%
5.70711026474888 0.817275747508306
};
\addplot [semithick, color0, opacity=0, mark=*, mark size=3, mark options={solid}, only marks]
table {%
5.49306144334055 0.732510288065844
};
\draw (axis cs:7.26822302115957,0.583682008368201) node[
  scale=1.2,
  anchor=base west,
  text=black,
  rotate=0.0
]{BCS};
\draw (axis cs:4.36944785246702,0.759493670886076) node[
  scale=1.2,
  anchor=base west,
  text=black,
  rotate=0.0
]{Bel};
\draw (axis cs:6.85646198459459,0.594736842105263) node[
  scale=1.2,
  anchor=base west,
  text=black,
  rotate=0.0
]{Bul};
\draw (axis cs:5.97126183979046,0.323979591836735) node[
  scale=1.2,
  anchor=base west,
  text=black,
  rotate=0.0
]{CS};
\draw (axis cs:7.22766249872865,0.565722585330429) node[
  scale=1.2,
  anchor=base west,
  text=black,
  rotate=0.0
]{Cze};
\draw (axis cs:4.78749174278205,0.75) node[
  scale=1.2,
  anchor=base west,
  text=black,
  rotate=0.0
]{LSo};
\draw (axis cs:7.00033446027523,0.333637192342753) node[
  scale=1.2,
  anchor=base west,
  text=black,
  rotate=0.0
]{OCS};
\draw (axis cs:7.15617663748062,0.497659906396256) node[
  scale=1.2,
  anchor=base west,
  text=black,
  rotate=0.0
]{Pol};
\draw (axis cs:7.36010397298915,0.513358778625954) node[
  scale=1.2,
  anchor=base west,
  text=black,
  rotate=0.0
]{Rus};
\draw (axis cs:6.99484998583307,0.575618698441796) node[
  scale=1.2,
  anchor=base west,
  text=black,
  rotate=0.0
]{Slk};
\draw (axis cs:7.28756064030972,0.536935704514364) node[
  scale=1.2,
  anchor=base west,
  text=black,
  rotate=0.0
]{Sln};
\draw (axis cs:5.70711026474888,0.817275747508306) node[
  scale=1.2,
  anchor=base west,
  text=black,
  rotate=0.0
]{Ukr};
\draw (axis cs:5.49306144334055,0.732510288065844) node[
  scale=1.2,
  anchor=base west,
  text=black,
  rotate=0.0
]{USo};
\end{axis}

\end{tikzpicture}
    }
    \end{minipage}
    \hspace{.03\linewidth}
    \begin{minipage}[t]{.3\linewidth}
    \adjustbox{width=\linewidth}{
\begin{tikzpicture}

\definecolor{color0}{rgb}{0.12156862745098,0.466666666666667,0.705882352941177}

\begin{axis}[
tick align=outside,
tick pos=left,
x grid style={white!69.0196078431373!black},
xmin=4.21991504644092, xmax=7.50963677901526,
xtick style={color=black},
y grid style={white!69.0196078431373!black},
ymin=0.306021594684385, ymax=0.869464878974846,
ytick style={color=black}
]
\addplot [semithick, color0, opacity=0, mark=*, mark size=3, mark options={solid}, only marks]
table {%
7.26822302115957 0.615062761506276
};
\addplot [semithick, color0, opacity=0, mark=*, mark size=3, mark options={solid}, only marks]
table {%
4.36944785246702 0.746835443037975
};
\addplot [semithick, color0, opacity=0, mark=*, mark size=3, mark options={solid}, only marks]
table {%
6.85646198459459 0.596842105263158
};
\addplot [semithick, color0, opacity=0, mark=*, mark size=3, mark options={solid}, only marks]
table {%
5.97126183979046 0.331632653061224
};
\addplot [semithick, color0, opacity=0, mark=*, mark size=3, mark options={solid}, only marks]
table {%
7.22766249872865 0.595497458242556
};
\addplot [semithick, color0, opacity=0, mark=*, mark size=3, mark options={solid}, only marks]
table {%
4.78749174278205 0.766666666666667
};
\addplot [semithick, color0, opacity=0, mark=*, mark size=3, mark options={solid}, only marks]
table {%
7.00033446027523 0.360072926162261
};
\addplot [semithick, color0, opacity=0, mark=*, mark size=3, mark options={solid}, only marks]
table {%
7.15617663748062 0.514040561622465
};
\addplot [semithick, color0, opacity=0, mark=*, mark size=3, mark options={solid}, only marks]
table {%
7.36010397298915 0.529898218829516
};
\addplot [semithick, color0, opacity=0, mark=*, mark size=3, mark options={solid}, only marks]
table {%
6.99484998583307 0.603116406966086
};
\addplot [semithick, color0, opacity=0, mark=*, mark size=3, mark options={solid}, only marks]
table {%
7.28756064030972 0.567715458276334
};
\addplot [semithick, color0, opacity=0, mark=*, mark size=3, mark options={solid}, only marks]
table {%
5.70711026474888 0.843853820598007
};
\addplot [semithick, color0, opacity=0, mark=*, mark size=3, mark options={solid}, only marks]
table {%
5.49306144334055 0.806584362139918
};
\draw (axis cs:7.26822302115957,0.615062761506276) node[
  scale=1.2,
  anchor=base west,
  text=black,
  rotate=0.0
]{BCS};
\draw (axis cs:4.36944785246702,0.746835443037975) node[
  scale=1.2,
  anchor=base west,
  text=black,
  rotate=0.0
]{Bel};
\draw (axis cs:6.85646198459459,0.596842105263158) node[
  scale=1.2,
  anchor=base west,
  text=black,
  rotate=0.0
]{Bul};
\draw (axis cs:5.97126183979046,0.331632653061224) node[
  scale=1.2,
  anchor=base west,
  text=black,
  rotate=0.0
]{CS};
\draw (axis cs:7.22766249872865,0.595497458242556) node[
  scale=1.2,
  anchor=base west,
  text=black,
  rotate=0.0
]{Cze};
\draw (axis cs:4.78749174278205,0.766666666666667) node[
  scale=1.2,
  anchor=base west,
  text=black,
  rotate=0.0
]{LSo};
\draw (axis cs:7.00033446027523,0.360072926162261) node[
  scale=1.2,
  anchor=base west,
  text=black,
  rotate=0.0
]{OCS};
\draw (axis cs:7.15617663748062,0.514040561622465) node[
  scale=1.2,
  anchor=base west,
  text=black,
  rotate=0.0
]{Pol};
\draw (axis cs:7.36010397298915,0.529898218829516) node[
  scale=1.2,
  anchor=base west,
  text=black,
  rotate=0.0
]{Rus};
\draw (axis cs:6.99484998583307,0.603116406966086) node[
  scale=1.2,
  anchor=base west,
  text=black,
  rotate=0.0
]{Slk};
\draw (axis cs:7.28756064030972,0.567715458276334) node[
  scale=1.2,
  anchor=base west,
  text=black,
  rotate=0.0
]{Sln};
\draw (axis cs:5.70711026474888,0.843853820598007) node[
  scale=1.2,
  anchor=base west,
  text=black,
  rotate=0.0
]{Ukr};
\draw (axis cs:5.49306144334055,0.806584362139918) node[
  scale=1.2,
  anchor=base west,
  text=black,
  rotate=0.0
]{USo};
\end{axis}

\end{tikzpicture}
    }
    \end{minipage}
    \hspace{.03\linewidth}
    \begin{minipage}[t]{.3\linewidth}
    \adjustbox{width=\linewidth}{
\begin{tikzpicture}

\definecolor{color0}{rgb}{0.12156862745098,0.466666666666667,0.705882352941177}

\begin{axis}[
tick align=outside,
tick pos=left,
x grid style={white!69.0196078431373!black},
xmin=4.21991504644092, xmax=7.50963677901526,
xtick style={color=black},
y grid style={white!69.0196078431373!black},
ymin=0.268521594684385, ymax=0.87125059326056,
ytick style={color=black}
]
\addplot [semithick, color0, opacity=0, mark=*, mark size=3, mark options={solid}, only marks]
table {%
7.26822302115957 0.587866108786611
};
\addplot [semithick, color0, opacity=0, mark=*, mark size=3, mark options={solid}, only marks]
table {%
4.36944785246702 0.721518987341772
};
\addplot [semithick, color0, opacity=0, mark=*, mark size=3, mark options={solid}, only marks]
table {%
6.85646198459459 0.574736842105263
};
\addplot [semithick, color0, opacity=0, mark=*, mark size=3, mark options={solid}, only marks]
table {%
5.97126183979046 0.295918367346939
};
\addplot [semithick, color0, opacity=0, mark=*, mark size=3, mark options={solid}, only marks]
table {%
7.22766249872865 0.566448801742919
};
\addplot [semithick, color0, opacity=0, mark=*, mark size=3, mark options={solid}, only marks]
table {%
4.78749174278205 0.758333333333333
};
\addplot [semithick, color0, opacity=0, mark=*, mark size=3, mark options={solid}, only marks]
table {%
7.00033446027523 0.315405651777575
};
\addplot [semithick, color0, opacity=0, mark=*, mark size=3, mark options={solid}, only marks]
table {%
7.15617663748062 0.494539781591264
};
\addplot [semithick, color0, opacity=0, mark=*, mark size=3, mark options={solid}, only marks]
table {%
7.36010397298915 0.495547073791349
};
\addplot [semithick, color0, opacity=0, mark=*, mark size=3, mark options={solid}, only marks]
table {%
6.99484998583307 0.570119156736939
};
\addplot [semithick, color0, opacity=0, mark=*, mark size=3, mark options={solid}, only marks]
table {%
7.28756064030972 0.545827633378933
};
\addplot [semithick, color0, opacity=0, mark=*, mark size=3, mark options={solid}, only marks]
table {%
5.70711026474888 0.843853820598007
};
\addplot [semithick, color0, opacity=0, mark=*, mark size=3, mark options={solid}, only marks]
table {%
5.49306144334055 0.724279835390947
};
\draw (axis cs:7.26822302115957,0.587866108786611) node[
  scale=1.2,
  anchor=base west,
  text=black,
  rotate=0.0
]{BCS};
\draw (axis cs:4.36944785246702,0.721518987341772) node[
  scale=1.2,
  anchor=base west,
  text=black,
  rotate=0.0
]{Bel};
\draw (axis cs:6.85646198459459,0.574736842105263) node[
  scale=1.2,
  anchor=base west,
  text=black,
  rotate=0.0
]{Bul};
\draw (axis cs:5.97126183979046,0.295918367346939) node[
  scale=1.2,
  anchor=base west,
  text=black,
  rotate=0.0
]{CS};
\draw (axis cs:7.22766249872865,0.566448801742919) node[
  scale=1.2,
  anchor=base west,
  text=black,
  rotate=0.0
]{Cze};
\draw (axis cs:4.78749174278205,0.758333333333333) node[
  scale=1.2,
  anchor=base west,
  text=black,
  rotate=0.0
]{LSo};
\draw (axis cs:7.00033446027523,0.315405651777575) node[
  scale=1.2,
  anchor=base west,
  text=black,
  rotate=0.0
]{OCS};
\draw (axis cs:7.15617663748062,0.494539781591264) node[
  scale=1.2,
  anchor=base west,
  text=black,
  rotate=0.0
]{Pol};
\draw (axis cs:7.36010397298915,0.495547073791349) node[
  scale=1.2,
  anchor=base west,
  text=black,
  rotate=0.0
]{Rus};
\draw (axis cs:6.99484998583307,0.570119156736939) node[
  scale=1.2,
  anchor=base west,
  text=black,
  rotate=0.0
]{Slk};
\draw (axis cs:7.28756064030972,0.545827633378933) node[
  scale=1.2,
  anchor=base west,
  text=black,
  rotate=0.0
]{Sln};
\draw (axis cs:5.70711026474888,0.843853820598007) node[
  scale=1.2,
  anchor=base west,
  text=black,
  rotate=0.0
]{Ukr};
\draw (axis cs:5.49306144334055,0.724279835390947) node[
  scale=1.2,
  anchor=base west,
  text=black,
  rotate=0.0
]{USo};
\end{axis}

\end{tikzpicture}
    }
    \end{minipage}

    \caption{PER (top) and WER (top) values (y axis) plotted by the log number of training examples for each language (x axis), for Dense, Sigmoid and ST models (left to right)}
    \label{fig:error}
\end{figure*}
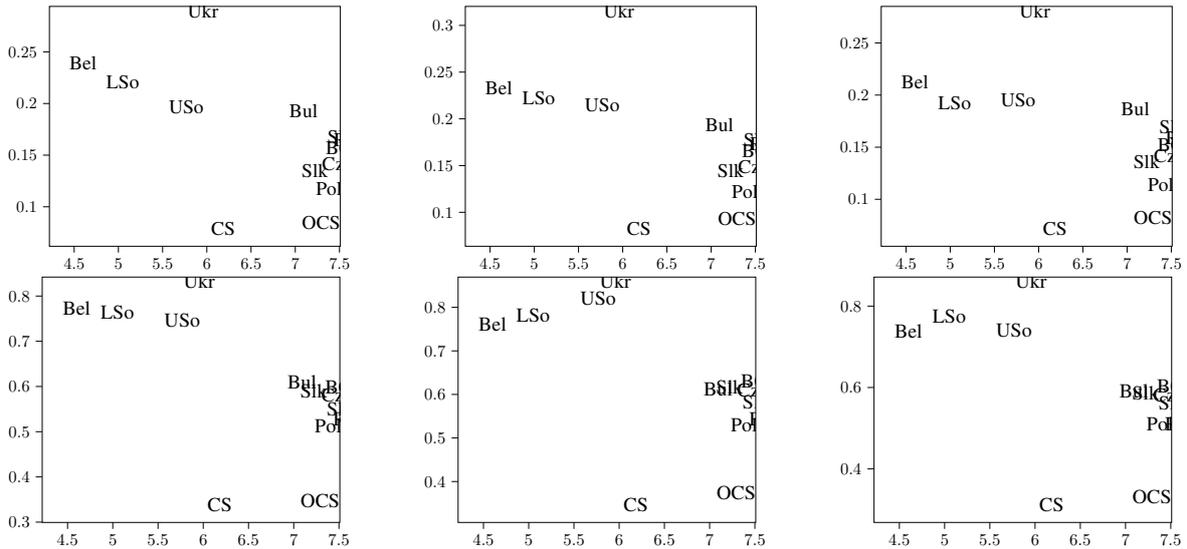

We assess the accuracy of each model by generating held-out forms on the basis of the language ID of the form and the Proto-Slavic etymon from which the form descends, greedily decoding on the basis of the trained model. 
We measure accuracy in terms of word error rate (WER), 
which gives the proportion of incorrectly generated forms, 
and the phoneme error rate (PER), which we define as the Levenshtein edit distance between generated and ground truth strings divided by the length of the longer form. Accuracy measures are found in Table \ref{tab:acc}. The ST model shows the best performance, followed by the Dense and Sigmoid models. 
Figure \ref{fig:error} shows WER and PER for each model plotted by the log number of training examples for each language in our data set. There is at best a very weak negative correlation between error rates and training example frequencies; the worst performance seems to be restricted to four languages (Belarusian, Lower Sorbian, Ukrainian, and Upper Sorbian), which vary in training data frequency, but in our impression posed the most difficulties for phonetic conversion. 
Old Church Slavic and Church Slavic show the highest accuracy; forms in these languages tend to be close to their Proto-Slavic ancestral forms,\footnote{
Note that according to the common practice in etymological dictionaries, OCS snd CS forms are given in a normalized form not reflecting regional differences.
}
and were straightforward to convert to IPA. 


\begin{table}
    \centering
    \begin{tabular}{|l|c|c|}
    \hline
     & WER & PER \\
    \hline
    Dense   & 0.535 & 0.143 \\
    Sigmoid & 0.559 & 0.151 \\
    ST      & \textbf{0.530} & \textbf{0.140} \\
    \hline
    \end{tabular}
    \caption{Mean word error and phoneme error rate for each model}
    \label{tab:acc}
\end{table}

\subsection{Error analysis}








\subsubsection{Quantitative error analysis}

We wish to obtain a fuller picture of the errors made by our models, and in particular, whether different models produce different types of errors. 
We analyze errors according to a taxonomy inspired by \citet{gorman2019weird}. 
At a high level, errors can be divided according to whether they stem from mistakes in the data or are a result of model idiosyncrasies. 
Errors in the data (target errors) largely consist of 
morphologically non-congruent etymon reflex pairs that we were unable to detect {\it a priori}: for instance, the Slavic etymon {*d\v{\i}lti} `to hollow, chisel' is paired with reflexes such as Czech {\it dlbsti}, which contains the cluster {\it -bs-} due to analogical influence; similarly, the etymon {*majati} `wave, beckon'  (inf.) is paired with OCS {\it namaiaax\k{o}} (3pl impf.). 
Additionally, there exists the possibility of doublet reflexes in contemporary Slavic languages due to dialect borrowing (free variation errors), e.g., Russian \textit{{\'o}blako} from Church Slavic \citep{vasmer1953}. 
Incorrect phonetic conversion is another source of errors of this type. 

In terms of linguistic errors that are not direct artifacts of our data set, we are interested in the degree to which the models' behavior results in a specific set of error pattern types. 
We wish to measure the extent to which models introduce errors when decoding forms in a given language due to overgeneralization on the basis of forms seen in the training data for the {\sc same language}. 
For instance, all models fail to learn the Upper Sorbian development {*pr} $>$ {\IPA [pS]}, erroneously generalizing the change {*r} $>$ {\IPA [\;R]} to an incorrect environment (e.g., PSl {*pr\k{e}sti} `spin' $>$ {\IPA ["p\;R\textsuperscript{j}as\t{tS}]}, expected {\IPA ["pSas\t{tS}]}). 
Additionally, because our model leverages global information shared across languages along with language-specific information, errors in one language involving the application of a sound law from a {\sc different Slavic language} are a potential concern. 
For example, the Sigmoid model generates the erroneous BCS reflex {\IPA [l\v{e}\t{tC}{\ae}ti]} `to fly' ($<$ PSl {*let\v{e}ti}, expected {\IPA [l\v{e}tjeti]}); 
{\IPA [\ae]} is attested only in OCS, CS, Russian, and Slovak.  
Of additional interest are errors where the model produces a rule that is unattested across the data set, and hence {\sc unmotivated} by the data. 
For instance, the Sigmoid model generates BCS {\IPA [pp\v{e}:ta]} ($<$ PSl {*p\k{e}t\`a} `heel', expected {\IPA [p\v{e}:ta]}); word-initial {\IPA [pp-]} is unattested in our data set, and the origin of this error is unclear. 

We quantitatively assess the issues enumerated above in the manner described below. To assess the prevalence of target errors, we measure the extent to which models agree in terms of the data points for which poor performance is exhibited. 
We take this agreement as a proxy for errors in the data; if the same data points cause problems across models, this poor performance may be an artifact of morphological mismatches in the data or fewer examples in the training data than needed to learn the patterns for the data points in question. The agreement matrix in Table \ref{tab:word_errors} shows that agreement levels are quite high, indicating that some errors may be due to artifacts of the data used. 

To gain an overview of the error types made by the model, we use the attention mechanism of the trained models to obtain alignments between all Proto-Slavic etyma and attested reflexes as well as between Proto-Slavic etyma and erroneously produced reflexes. We extract sound changes operating between Proto-Slavic and daughter languages from these alignments 
(e.g., PSl {*o} $>$ Slovak {\IPA O}), which indicate whether a given edit is attested in a language (irrespective of conditioning environment). We automatically annotate each erroneous edit according to whether it is attested in the same language as the decoded form in which it occurs (same language), if not, whether it is attested in another Slavic language (other language), or finally, if it is not attested in any Slavic language (unmotivated). 
Table \ref{tab:error} shows proportions of these error types produced by each model; the Sigmoid and ST models produce more other-language and unmotivated errors than the Dense model.

\begin{table}
    \centering
    \begin{tabular}{|l|c|c|c|}
    \hline
        Model & Dense & Sigmoid & ST \\
        \hline
        Dense & --- & 0.789 & 0.812 \\
        Sigmoid & 0.824 & --- & 0.827\\
        ST & 0.803 & 0.783 & --- \\
        \hline
    \end{tabular}
    \caption{Proportion of word errors produced by each model (rows) shared with other models (columns)}
    \label{tab:word_errors}
\end{table}

\begin{table}
    \centering
    \begin{tabular}{|l|c|c|c|}
    \hline
    Model & SL & OL & U\\
    \hline
    Dense & 0.551 & 0.105 & 0.342\\
    Sigmoid & 0.593 & 0.101 & 0.305\\
    ST & 0.617 & 0.101 & 0.281\\
    \hline
    \end{tabular}
    \caption{Proportion of errors produced by models that are present in the same language (SL), other languages (OL), or are unmotivated (U)}
    \label{tab:error}
\end{table}

\subsubsection{Qualitative error analysis}

We present results of a detailed error analysis involving 422 forms spanning all languages in the data set where at least one of the three models 
produced an error. 
Roughly 15\% of the forms surveyed contain some sort of morphological mismatch; many of these are trivial one-off analogical idiosyncrasies. 
In some cases, loanwords unmarked in the dictionary can be detected (cf.\ the example of Russian {\it \'oblako} mentioned above). 


Annotated error types that occur more than once across all models include incorrect accent type (Dense: 18, Sigmoid: 12, ST: 10), accent misplacement (Dense: 40, Sigmoid: 40, ST: 32), consonant mismatches (Dense: 139, Sigmoid: 161, ST: 149), vowel quality mismatches (Dense: 192, Sigmoid: 219, ST: 195), vowel length mismatches (Dense: 35, Sigmoid: 47, ST: 31), and general segmental mismatches involving the erroneous substitution of a vowel for a consonant, or vice versa (Dense: 85, Sigmoid: 81, ST: 68). 
The ST model's overall higher performance bears out the larger-scale analysis of errors presented in the previous section. 

Our manual error analysis was carried out by a single specialist; future research will involve more detailed error analyses carried out by multiple specialists in order to gauge inter-annotator reliability.

\subsection{Genetic signal in embeddings}

\begin{figure*}
\centering
\adjustbox{width=.9\linewidth}{
\input{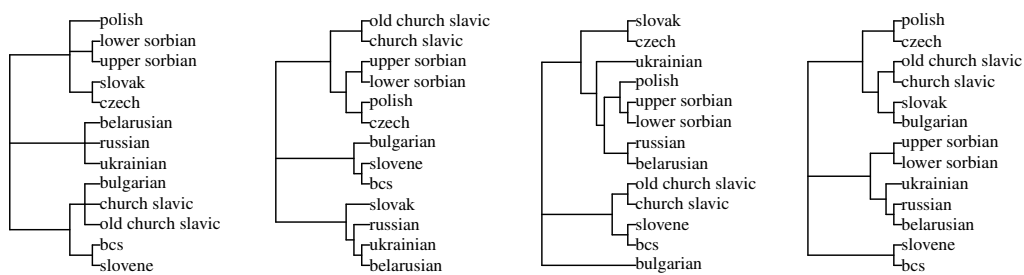}
}
\caption{Reference phylogeny of Slavic languages and neighbor-joined trees from embeddings for Dense, Sigmoid, and Straight-Through models (left to right)}
\label{fig:gen}
\end{figure*}

We wish to measure the degree to which the language-level embeddings learned by each model reflect received wisdom regarding the dialectal makeup of the Slavic languages. As stated previously, languages employed in this paper are 
traditionally 
divided among East, South and West Slavic groups. 
To assess the signal contained by the embeddings, we generated trees from cosine distances between pairs of language embeddings learned by each model using neighbor joining \citep[NJ, ][]{SaitouNei1987} as implemented in the R package {\tt ape} \cite{paradis2019package}. 
These trees can be found in Figure \ref{fig:gen} alongside a reference topology from Glottolog \citep{Glottolog}. 
The Sigmoid model's embeddings show the highest agreement with the Glottolog tree; the main discrepancies found are the placement of Bulgarian outside of South Slavic, as well as the placement of the Lechitic languages Polish, Upper Sorbian and Lower Sorbian within East Slavic. 
The ST embeddings show mixed performance; certain West and South Slavic languages are grouped correctly, but a large number of taxa are misplaced. 
We used the R package {\tt Quartet} \citep{Smith2019} to measure the generalized quartet distance \citep{pompei2011accuracy} between the reference tree and the trees constructed from the embeddings, equal to the number of four-taxon groups resolved differently across the two trees, divided by the number of resolved four-taxon groups found in the reference tree; lower values indicate greater agreement 
(Dense: 0.322, Sigmoid: {\bf 0.247}, ST: 0.368). 
It is possible that the ST model shows low agreement with the reference phylogeny but high accuracy because it has succeeded in detecting areal features that conflict with the traditional tripartite subgrouping. Further investigation into the treelikeness of each network \citep{wichmann2011correlates} is needed in order to properly address this issue.  


\subsection{Interpretation of embeddings}

\begin{figure}
\centering
\includegraphics[width=.9\linewidth]{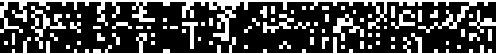}
\caption{Active dimensions (white cells) in ST language embeddings}
\label{STheatmap}
\end{figure}





A common goal of neural modeling with discrete latent variables is to learn sparse interpretable features. Ideally, activating or deactivating a single binary latent variable should correlate with the presence or absence of a meaningful feature in the model's output. Inducing the level of sparsity needed to generate such latent variables is an ongoing issue in the deep learning literature \cite{singh2017structured}. 
Our models have not learned meaningful features in the sense that turning a single variable ``on'' or ``off'' can produce a meaningful feature of Slavic dialectology (e.g., the presence of liquid metathesis/pleophony); these processes appear to be distributed across multiple latent binary variables. 

As shown in Figure \ref{STheatmap}, language-level straight-through embeddings are far from sparse; of the 128 embedding dimensions, only 1 is inactive across languages. Individual language embeddings contain between 32 and 61 active dimensions. Preliminary attempts to turn individual dimensions ``on'' and feed the latent representation to the encoder-decoder along with a Proto-Slavic input do not produce interpretable or coherent results; it appears to be the case that it is not individual dimensions, but interactions between them, that influence the behavior of the decoder. 

\subsubsection{Nearest neighbors}

Feeding all possible $2^{128}$ combinations of embedding values to the model is computationally infeasible, though it might allow us to discover which feature combinations are responsible for certain types of behavior of the encoder-decoder. In order to gain a better understanding of the behavior of these dimensions individually and as a group, we explore the {\sc nearest neighbors} in embedding space of reflexes for languages in our data set by altering the values of each variable in our language embeddings, and feeding these altered embeddings to the model architecture along with a Proto-Slavic etymon and observing the set of resulting outputs. 
Specifically, we take language-level embeddings and alter each of the embedding's 128 dimensions.

By feeding these nearest neighbors to the encoder-decoder along with a Proto-Slavic etymon, it is possible to see how perturbations of an embedding result in different outputs from the expected contemporary Slavic reflex. In general, different perturbations often result in the same output form, indicating that the embedding space is perhaps less sparse than necessary, and more compact representations can be learned without losing information. 
To give a concrete example of this phenomenon, the nearest neighbors of Polish 
{\IPA ["m1dwO]} `soap, lather' ($<$ PSl {*m\`ydlo}) in embedding space yield only thirteen unique forms ({\IPA ["milo]}, 
{\IPA ["midlO]}, 
{\IPA [m\v{\i}:dlO]}, 
{\IPA ["m1dwO]}, 
{\IPA [m\v{\i}:lo]}, 
{\IPA ["m1dlO]}, 
{\IPA [mIll5]}, 
{\IPA [mIllO]}, 
{\IPA [m\v{\i}lO]}, 
{\IPA [mIlO]}, 
{\IPA ["milO]}, 
{\IPA [m\v{\i}:lO]}); 
interestingly, there is no evidence for the otherwise naturalistic and plausible sound change {*dl} $>$ {\IPA [ll]} in our data set. 



Based on a qualitative appraisal of these nearest neighbors, it does not appear to be the case that the ST model has learned to entirely disentangle orthogonal developments in historical phonology. 
The unique nearest neighbors of Russian {\IPA [m@l5"ko]} `milk' ($<$ PSl {*melk\`o}), 
{\IPA [mlEkO]}, 
{\IPA ["mlEko]}, 
{\IPA [ml\v{e}:kO]}, 
{\IPA [mlI"ko]}, 
{\IPA [mlě:ko]}, 
{\IPA ["mlEkO]}, 
{\IPA [m@l5"ko]} and 
{\IPA [mlE"kO]}, 
appear to show that our model learns patterns of pleophony/liquid metathesis and vowel change jointly, rather than learning disentangled abstractions (though interestingly, the same word in Polish has the neighbors {\IPA ["m\textsuperscript{j}IlkO]} and {\IPA [ml\textsuperscript{j}I"kO]}, showing metathesis independent of vowel quality). 
It is not clear, however, that this behavior goes against the received wisdom of Slavic linguistics; the operation of liquid metathesis or pleophony among Slavic languages is generally thought to be a change that has an early common origin but developed in different dialect-specific directions \citep{shevelov1964}. 
Ultimately, this architecture shows the potential to generate typologically meaningful (i.e., naturalistic) but also novel representations of hypothetical Slavic reflexes.


\subsubsection{Sampling from the latent space}
An issue that arises in the use of latent variable models, particularly in the context of linguistic typology, concerns the coherence of the representations that they learn. If we randomly traverse our models' latent variable space or interpolate between representations, how likely are we to encounter a plausible unattested sister language of the languages attested in our data set? 
We briefly explore this question by randomly sampling 100 embeddings from variously parameterized distributions and feeding them to our models, along with a set of 100 randomly chosen Proto-Slavic etyma. For each etymon, we feed zero-mean Gaussian samples with standard deviation $\sigma \in \{.01,.1,1,10\}$ to the Dense model; symmetric Beta samples with shape parameters $\alpha = \beta \in \{.01,.1,1,10\}$ to the Sigmoid model; and Binomial samples with probability $p \in \{.2,.4,.6,.8\}$ to the ST model (all samples have the same dimension as our learned embeddings). 
Qualitatively speaking, output forms randomly generated by the ST model are consistently well formed and coherent across parameterization regimes. Conversely, when $\sigma$ is greater than $.01$ (roughly equivalent to the empirical standard deviation of the learned embeddings), the Dense model often generates unrealistic strings (e.g., {\IPA [b{\textbarl}{\textbarl}":{\textbarl}{\textbarl}@]}), and when $\sigma$ is very small, forms are coherent but there is virtually no variation; for the Sigmoid model, the strings become more realistic looking as $\alpha = \beta$ increases (the majority of values for the learned Sigmoid embeddings are close to $.5$). To highlight a related discrepancy, we observe the average number of unique outputs generated by each regime in each model (Dense: 3.21, 24.02, 93.08, 61.3; Sigmoid: 96.23, 94.14, 67.39 22.74; ST: 21.3, 24.3, 24.06, 20.1); the quantity of unique outputs stays constant across all regimes for the ST model, along with their quality. 


Additionally, we wish to explore the extent to which samples from latent variable space generate realistic sound changes and plausible sound patterns. 
While certain diachronic trajectories can lead to the emergence of ``crazy'' rules  \citep{BachHarms1972,buckley2000naturalness} and unnatural phonotactic restrictions \citep{beguvs2017lexicon}, we might expect the relatively infrequent nature of 
these phenomena 
to somehow be captured by the behavior of models like the ST model. 
To address this question, we feed sets of hypothetical well-formed Proto-Slavic phonological neighbors (generated by taking 12 etyma from our data set and generating echo-forms to create a cohort of forms differing according to initial {*p-/t-/k-/b-/d-/g-}; we exclude hypothetical forms with velar-front vowel sequences, which would have been affected by palatalization) 
to the ST model, randomly sampling binary latent embeddings from the binomial distribution with probabilities  $\{.2,.4,.6,,8\}$. 
For each probability regime, we 
attempt to 
evaulate the relative frequency of unnatural sound patterns 
displayed by these hypothetical forms' descendants; if our model embodies not only plausible but probable behavior, we predict that these etymological phonological neighbors, which differ only according to the word-initial consonant, should frequently yield similar echo-forms, 
and that other patterns may arise less frequently. 
For each pair of outputs within each cohort (with stress marking removed), we divide the number of agreeing final segments by the mean of the two strings' lengths, and report the proportion of pairs for which this value is greater than .5 (indicating greater agreement); these values are 0.427, 0.546, 0.574, and 0.526 for each respective probability regime, indicating that generated outputs tend not to be very echo-like. 
To exemplify, a representative sample output for {*\{p,t,b,d\}\u{\i}rt\u{\i}} comprises the forms {\IPA ["p\s{r}\t{tC}]}, 
{\IPA ["t\s{r}t]}, 
{\IPA ["b\s{r}\t{ts}]}, 
{\IPA ["d\s{r}t]}. 
While long-distance assimilatory and dissimilatory processes operating between the 
left and right word edge are not unknown cross-linguistically, 
we believe that changes where 
differences in word-initial segments trigger divergent word-final reflexes 
should be rare, rather than typical. 
Further refinement of metrics designed to assess the validity of output patterns 
is much needed.\footnote{Indeed, taking the proportion of agreeing final segments as a measure of naturalness would classify changes resulting from certain types of tonogenesis to be unnatural, e.g., {*pa, *ba} $>$ Vietnamese {\it pa}, {\it p\`a} \citep{Haudricourt1954}, since dissimilarity in initial consonants often leads to dissimilarity at the right word edge.}


From this small and rather premature investigation, it appears that the latent variable design space represented by the ST model generates 
coherent, realistic-looking output, but the frequency distributions of patterns in its output may not reflect cross-linguistic frequency distributions. 
A more in-depth analysis along these lines is outside the scope of this paper, and methods seeking to derive typological generalizations should include data from multiple families; at the same time, 
the issues raised here potentially bear on our understanding of the diachronic basis of
synchronic patterns in phonology. 


%


\section{Discussion and outlook}

This paper investigated the performance of multiple neural models in capturing patterns of sound change across Slavic languages. We found that a model with binarized straight-through language-level embeddings outperformed other models in terms of accuracy, and shows great potential for learning coherent and interpretable information regarding sound change. 
We found that the discrete features learned by our model appear for the most part to correspond to meaningful, realistic variation in sound patterns, though representations are not particularly sparse. 
Additionally, randomly sampling from discrete latent space tended to consistently generate coherent output; the preliminary attempts that we made to assess the likelihood of observing these samples in naturalistic contexts can be expanded considerably. 


We used straight-through embeddings as a low-cost alternative to more involved means of training discrete latent variables.  In the immediate future, we plan to extend our approach to make use of variational approaches, the flexibility of which may help in inducing sparsity in order to learn more meaningful, realistic representations (there is additionally room for exploration of simpler approaches that we did not make use of in this paper, such as dropout regularization); however, since our encoder-decoder model is different from the autoencoding models used in previous work, directly extending these methods presents a challenge that require considerable experimentation to overcome (an early attempt to adopt the IBP prior of \citealt{singh2017structured} was unsuccessful, as the monotonically decreasing prior probabilities rarely yielded non-zero values; thus far, attempts to weight the KL divergence term have not yielded success). Nevertheless, as low-variance, low-bias techniques for inferring discrete variables in neural models progress, we believe that they will be an increasingly valuable means of capturing meaningful, interpretable features in multilingual neural tasks like this paper's.





\bibliographystyle{acl_natbib}
\bibliography{coling2020}

\end{document}